%% file: main.tex
\documentclass[10pt,twocolumn,letterpaper]{article}

\usepackage[pagenumbers]{main}   % arXiv/Clean (page numbers, no specific identifiers)

\usepackage{multirow}    % Required for \multirow command
\usepackage{amssymb}     % Required for \checkmark (\blackcheck)
\usepackage{xcolor}      % Required for \textcolor (grayx)
\usepackage{booktabs}    % Required for \toprule/\midrule/\bottomrule
\usepackage{graphicx}


\definecolor{cvprblue}{rgb}{0.21,0.49,0.74}
\usepackage[pagebackref,breaklinks,colorlinks,allcolors=cvprblue]{hyperref}

\definecolor{cvprgreen}{rgb}{0.10, 0.52, 0.27}
\definecolor{cvprgrey}{rgb}{0.5, 0.52, 0.5}
\definecolor{darkblue}{RGB}{0,0,180} 
\definecolor{darkred}{RGB}{180,0,0} 
\definecolor{darkpurple}{RGB}{120,0,180} 
\definecolor{darkgreen}{RGB}{0,120,0} 
\definecolor{darkbrown}{RGB}{100,40,0} 
\definecolor{grey}{RGB}{128,128,128} 

\title{Leveraging Unlabeled Data from Unknown Sources via Dual-Path Guidance for Deepfake Face Detection}

\author{
    Zhiqiang Yang\textsuperscript{1,2}, 
    Renshuai Tao\textsuperscript{1,2}\thanks{Corresponding author.}, 
    Chunjie Zhang\textsuperscript{1,2},
    Guodong Yang\textsuperscript{3}, 
    Xiaolong Zheng\textsuperscript{3}, 
    Yao Zhao\textsuperscript{1,2} 
    \\[0.5em] 
    \textsuperscript{1}Institute of Information Science, Beijing Jiaotong University\\
    \textsuperscript{2}Visual Intelligence +X International Cooperation Joint Laboratory of MOE\\
    \textsuperscript{3}Institute of Automation, Chinese Academy of Sciences\\
    {\tt\small \{yzq1636, rstao, cjzhang,yzhao\}@bjtu.edu.cn, \{xiaolong.zheng, guodong.yang\}@ia.ac.cn}
}
\usepackage{bibentry}
\begin{document}
\maketitle
\input{0_abstract}    
\input{1_intro}

\input{2_related}
\input{3_method}

\input{4_experiments}
{
 \small
    \bibliographystyle{ieeenat_fullname}
    \bibliography{main}
}

\end{document}

%% file: 0_abstract.tex
\begin{abstract}
Existing deepfake detection methods heavily rely on static labeled datasets. However, with the proliferation of generative models, real-world scenarios are flooded with massive amounts of unlabeled fake face data from unknown sources. This presents a critical dilemma: detectors relying solely on existing data face generalization failure, while manual labeling for this new stream is infeasible due to the high realism of fakes. A more fundamental challenge is that, unlike typical unsupervised learning tasks where categories are clearly defined, real and fake faces share the same semantics, which leads to a decline in the performance of traditional unsupervised strategies. Therefore, there is an urgent need for a new paradigm designed specifically for this scenario to effectively utilize these unlabeled data. Accordingly, this paper proposes a dual-path guided network (DPGNet) to address two key challenges: (1) bridging the domain differences between faces generated by different generative models; and (2) utilizing unlabeled image samples. The method comprises two core modules: text-guided cross-domain alignment, which uses learnable cues to unify visual and textual embeddings into a domain-invariant feature space; and curriculum-driven pseudo-label generation, which dynamically utilizes unlabeled samples. Extensive experiments on multiple mainstream datasets show that DPGNet significantly outperforms existing techniques,, highlighting its effectiveness in addressing the challenges posed by the deepfakes using unlabeled data. \footnote{The code is in the supplementary material and will be open-sourced after publication.}
% To prevent catastrophic forgetting, we also facilitate knowledge connections between different domains through cross-domain feature enhancement.
\end{abstract}

%% file: 1_intro.tex
\vspace{-0.25in}
\section{Introduction}
\label{sec:intro}

The proliferation and accessibility of generative models \cite{zhou2023generative,yu2023reinforcement,zhang2023encrypted,zhang2024dibad,pan2024few,zhang2024few,liu2024learning} has led to an unprecedented surge in deepfakes, particularly in the form of face forgery. As a response, the dominant research paradigm \cite{huang2025sida, liu2024forgerygpt, xu2024fakeshield, wang2025mllm, yang2025heie,peng2025mllm,guo2025rethinking} has relied on supervised learning. These methods are trained on static, curated datasets which are constructed using a limited, known set of forgery techniques. While effective in-domain, this reliance on known generators creates a fundamental vulnerability and is increasingly misaligned with the current media ecosystem.

\begin{figure}[!t]
% \vspace{-0.1in}
\centerline{\includegraphics[width=0.46\textwidth]{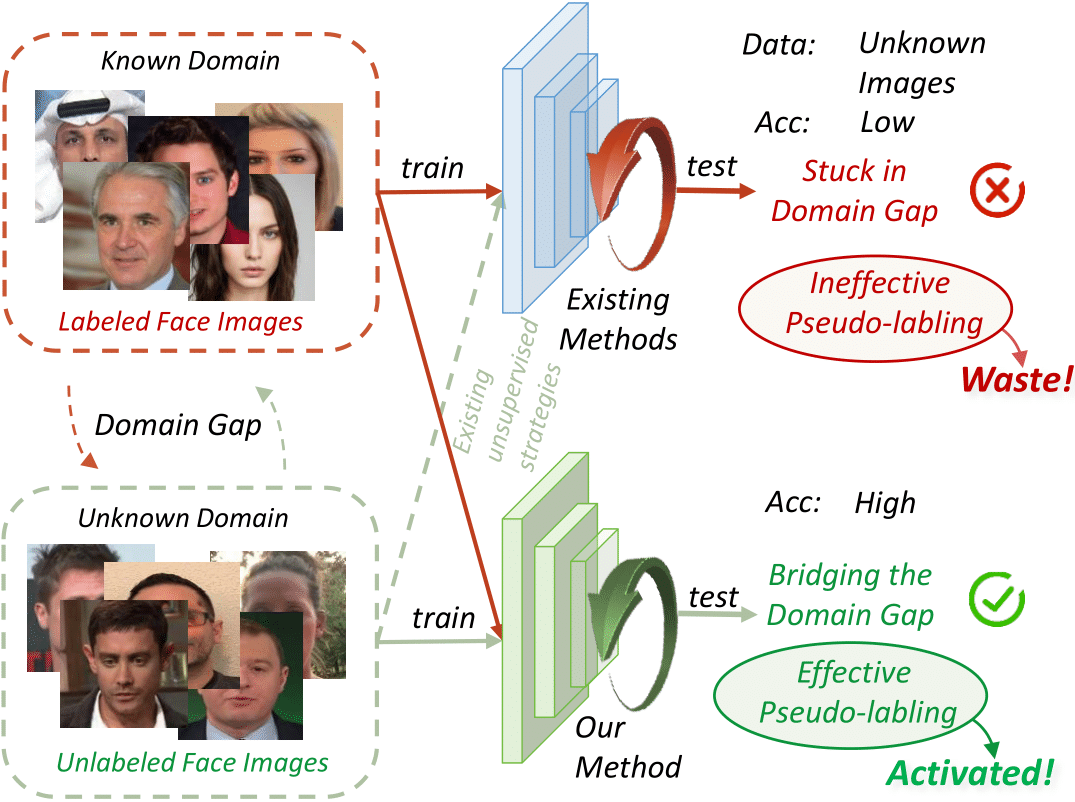}} %5、6、7
\vspace{-0.05in}
\caption{Comparison between traditional labeled data training and the new proposed unlabeled data training setting.}
\vspace{-0.15in}
\label{fig:12}
\end{figure}

To address this vulnerability, much existing research has focused on improving generalization \cite{laa-net,qiao2024fully,lin2024preserving,cfm,effort,ForensicsAdapter}, aiming to create detectors that are robust to unseen forgery types. However, we posit that this defense-only approach overlooks a crucial opportunity. The reality of online platforms is a massive, continuous stream of unlabeled content where the generative source is unknown. We believe this data stream, rather than being merely an obstacle, can be actively leveraged to enhance detection. This perspective, however, leads to a critical dilemma: the unknown source reality means the data is unlabeled. The seemingly straightforward alternative of manually labeling this unlabeled data is fundamentally impractical. The high realism of modern fakes makes any such attempt both time-consuming and unreliable. This highlights an urgent need for a new paradigm: \textbf{one that can effectively leverage this unlabeled and unknown-source data stream.}

\textbf{Can traditional unsupervised learning methods handle this task?} A key challenge in deepfake detection is that faces generated by different AI models closely mimic the distribution of real human faces and are often highly similar to each other. \textbf{Unlike typical unsupervised learning tasks, where semantic categories are well-separated, faces produced by generative models share many common features with real faces, leading to significant overlap.} This overlap complicates the task for traditional unsupervised methods, which rely on clearly defined categories to differentiate between real and fake. As a result, existing unsupervised learning~\cite{bai2024prompt,yu2025feature,zhuang2022uia,deng2025multi,zhang2025link,unsurpervised} approaches struggle to capture the subtle differences between real and fake faces, resulting in lower performance and reduced effectiveness in practical applications.
\

In this work, we introduce the \textbf{D}ual-\textbf{P}ath \textbf{G}uidance \textbf{Net}work(DPGNet), a novel framework designed to tackle the challenge introduced above. Unlike traditional methods that rely solely on labeled data, DPGNet combines two paths. First, we retain the original labeled data from traditional training settings as the source domain. The second path takes advantage of large-scale unlabeled data, often sourced from online social networks, which reflects the abundant real-world data available for training. DPGNet addresses two main challenges: (1) bridging the gap between labeled source data and the diverse, unlabeled data generated by different AI models, and (2) effectively utilizing large-scale unlabeled images. DPGNet consists of two key components: text-guided cross-domain alignment and curriculum-driven pseudo label generation. The first component uses learnable prompts to align visual and textual information into a shared, domain-independent feature space. This allows the model to better handle different types of deepfake faces while leveraging textual information. The second component mimics human learning by gradually incorporating and learning from more informative unlabeled samples. Through dynamic threshold supervision, it ensures the model focuses on the challenging samples. To prevent the loss of previously learned information, we also introduce the cross-domain feature enhancement. This ensures that the source domain's representation is bridged while adapting to new data.

We conduct extensive experiments across multiple datasets, including both cross-domain and cross-method evaluations, to evaluate the effectiveness of DPGNet. Our results show that the proposed method outperforms SoTA methods, achieving a significant improvement of detection in detection accuracy. These findings highlight the ability of our method to effectively leverage unlabeled data in real-world scenarios, overcoming the annotation challenges caused by the increasing realism of deepfakes and providing a scalable solution for face forgery detection. The main contributions are summarized as follows:

\begin{figure}[!t]
\vspace{-0.2in}
\centerline{\includegraphics[width=0.5\textwidth]{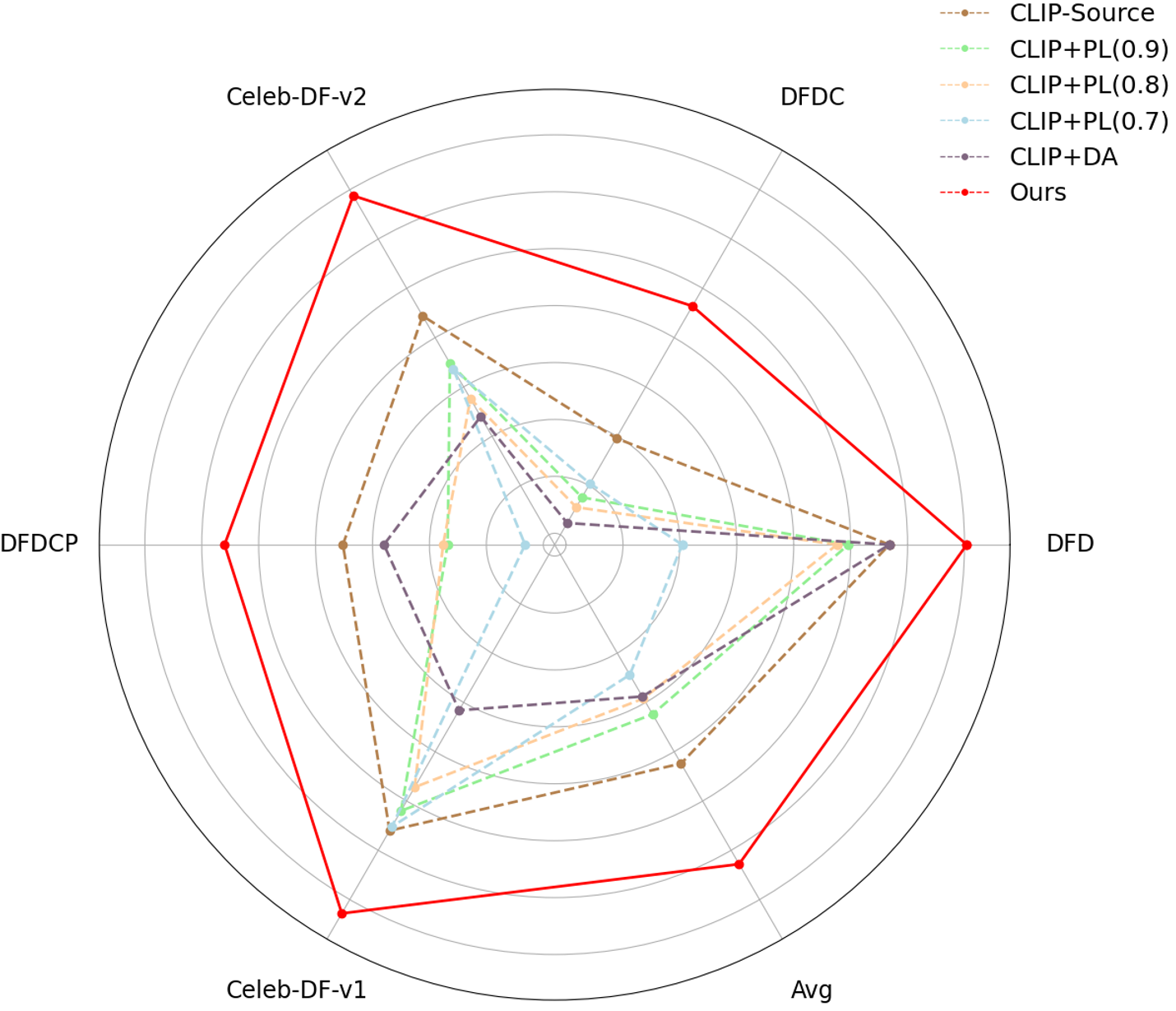}}
\caption{Comparison of methods leveraging unlabeled data. CLIP-Source is trained on FF++~\cite{ff++}. CLIP+PL(x) variants are fine-tuned using UCDDP pseudo-labels with confidence thresholds $\geq 0.9$, $0.8$, and $0.7$. CLIP+DA uses domain alignment with UCDDP ~\ref{tab:dataset_comparison}. See Section ~\ref{method-problem} for the details of the dataset.}
\vspace{-0.2in}
\label{fig:fig2}
\end{figure}

% We conducted extensive experiments on deepfake detection benchmarks and were surprised to find that our method can significantly improve cross-domain performance at very little unlabeled training cost.

\begin{itemize}
\item We are the first to propose a paradigm for effectively utilizing unlabeled data for face forgery detection, shifting from a generalization used only for defense to actively utilizing unlabeled data streams from unknown sources.
\item We propose DPGNet, a novel dual-path framework that synergizes text-guided domain alignment and curriculum-driven pseudo-labeling to effectively learn from both labeled and unlabeled data streams.
% \item We conduct comprehensive experiments across multiple datasets, demonstrating the effectiveness of our approach in deepfake detection and showing significant improvements over state-of-the-art methods.
\item Comprehensive experiments have validated the effectiveness of DPGNet in various settings on multiple popular datasets and demonstrated its superior performance.
% It is worth noting that our method can be integrated into other detectors and bring effective improvements.
\end{itemize}

%% file: 2_related.tex
\section{Related Work}
\label{sec:formatting}

Our work focuses on leveraging unlabeled data from unknown generators in face forgery detection to bridge the domain gap between source and target domains. We review related works on generalizable deepfake detection and traditional domain adaptation, highlight their limitations, and elaborate on our contributions.

\textbf{Generalizable deep fake image detection. } To address the emergence of new forgery methods, generalization has become a mainstream direction. Early works focused on mining general artifacts, such as frequency-domain discrepancies \cite{li2021frequency,luo2021generalizing,liu2021spatial}, or used reconstruction, decoupling, and distillation to learn shared fake features \cite{yan2023ucf,fu2025exploring,huang2023implicit,dong2023implicit,lsda}. In recent studies, the field has shifted to leveraging the powerful features of V-LMs like CLIP, using adapters \cite{ForensicsAdapter} or specialized fine-tuning \cite{effort,2025unlocking,guo2025rethinking}. Concurrently, MLLMs (Multimodal Large Language Models) are also being actively explored, with a growing number of methods \cite{huang2025sida, xu2024fakeshield, wang2025mllm, yang2025heie, guo2025rethinking} leveraging the strong zero-shot reasoning and generalization capabilities inherited from their large-scale pre-training to assess image.

\textbf{Unsupervised Domain Adaptation (UDA).} UDA aims to address the domain gap using unlabeled target data. Methods in this area range from traditional adversarial alignment \cite{long2017conditional,ganin2015unsupervised} to more recent techniques in knowledge transfer \cite{ma2025steady} and adaptive alignment \cite{he2025differential}. However, these methods perform poorly in deep face forgery detection, as the task fundamentally violates the core assumptions of UDA in two ways. First, the very definition of a {domain} is misaligned. In traditional UDA, domains have clear stylistic differences (e.g., {photos} vs. {paintings}). In forgery detection, however, different {domains} all appear as {faces}, rendering the inter-domain differences subtle and non-semantic. Second, the {categories} themselves are semantically inseparable. Unlike traditional classification tasks (e.g., \emph{cat} vs. \emph{dog}), real and fake faces share the identical high-level attribute of {face}. The distinction lies in subtle artifact traces, not in different semantic classes.

This dual challenge of non-semantic domains and semantically-identical categories invalidates the core premise of traditional feature alignment methods. To bypass this alignment failure, pseudo-labeling strategies \cite{unsurpervised} are often employed. However, high-confidence pseudo-labels often prioritize simple samples with low value, leading to error propagation and ignoring challenging cases that are critical for robust detection. Some parameter efficient fine-tuning (PEFT) methods \cite{PEFT}, such as LoRA~\cite{lora}, achieve domain adaptation by fine-tuning the visual base model, but there is a risk of distorting the pre-trained knowledge \cite{cai2019learning,tang2020unsupervised}, which may lead to low ranking \cite{effort} in the feature space.
Our method overcomes these limitations by integrating visual-language alignment and dynamic pseudo-labeling, effectively capturing various artifact patterns while preserving prior knowledge and ensuring robust generalization across domains.

\begin{figure*}[!t]
\vspace{-0.2in}
\centerline{\includegraphics[width=1.0\textwidth]{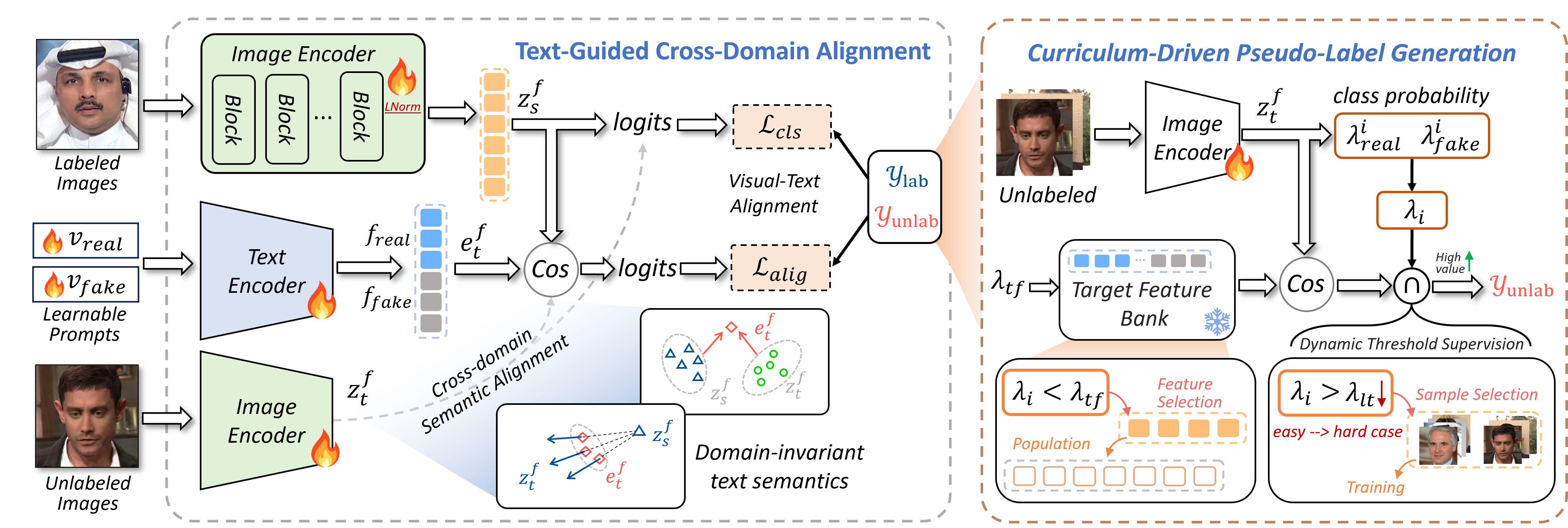}}
\caption{Framework overview of the proposed DPGNet, illustrating the overall architecture and the interaction between its two core modules: text-guided cross-domain alignment and curriculum-driven pseudo label generation.}
\vspace{-0.2in}
\label{fig:model}
\end{figure*} 

%% file: 3_method.tex
\section{Methodology}

% In this section, we elaborate on the DPGNet that we have introduced above. Firstly, we provide a problem definition of the unsupervised deepfake detection. Secondly, we categorize generalization gaps into three levels and provide a detailed explanation of their differences (see Section \ref{sec:3.2}). Following that, we provide an overview of our proposed method (see Section \ref{sec:3.3}), which includes the core Category-Specific Content Erasing Module (CSCE) and details about the Entire Training Procedure.

\subsection{Problem Definition}\label{method-problem}

The task involves a labeled source domain dataset $\mathcal{D}_s = \{(x_i^s, y_i^s)\}_{i=1}^{N_s}$ and an unlabeled target domain dataset $\mathcal{D}_t = \{x_i^t\}_{i=1}^{N_t}$ , where $x_s^i$ is an image and $y_s^i \in \{0, 1\}$ represents the label. Our goal is to solve the problem of covariate shift between $\mathcal{D}_s$ and $\mathcal{D}_u$, and effectively utilize a small number of target domain samples $\mathcal{D}_t$ extracted from multiple cross-domain datasets with different distributions, so that the model can generalize well to the full target domain data $\mathcal{D}_u = \{x_t^i \}_{i=1}^{N_u}$, where $N_t$ and $N_u$ represent the number of samples, and through our setting $N_t \ll N_u$.

To emulate the diverse forgery techniques encountered in real-world scenarios, we constructed two composite datasets, UCDDP and UDF40, by sampling a small subset of images from the training sets of multiple cross-domain datasets $\{\mathcal{D}_1, \mathcal{D}_2, \dots, \mathcal{D}_K\}$. This approach ensures diversity in unknown forgery methods and data distributions. Specifically, UCDDP encompasses images generated by various unknown forgery techniques, exhibiting significant distributional variations across samples. In contrast, UDF40 includes forgery methods distinct from those in the source domain while maintaining a consistent data distribution with the source domain. The task is formalized as minimizing the classification error on the target domain:
\begin{equation}
\min_f \mathbb{E}_{\mathbf{x}_u \in \mathcal{D}_u} \left[ \ell(f(x_u), y_u) \right]
\end{equation}
where $\ell$ denotes the classification loss function and $y_u$ represents the latent true label of the sample.

% 将z_0换为z_s
% \begin{figure*}[!t]
% \centerline{\includegraphics[width=1.0\textwidth]{CameraReady/LaTeX/figure/model-color2.png}}
% \caption{Method overview: It mainly includes the source domain training stage and the joint training stage, where the joint training stage includes the generation and utilization of pseudo labels, which is after the source domain training stage.}
% \label{fig:model}
% \end{figure*}    

\subsection{Framework Overview}
% As illustrated in Figure~\ref{fig:fig2}, we evaluate four approaches for leveraging unlabeled data: (1) a baseline excluding unlabeled data, (2) high-confidence pseudo-labeling, (3) domain alignment, and (4) our proposed method. Experiments reveal that conventional pseudo-labeling often prioritizes easily classified samples, typically simplistic pseudo-samples with limited generalization potential. Moreover, the domain gap causes visual embeddings in $\mathcal{D}_t$ to diverge from the source domain's trained model, leading to unreliable feature alignment.
We conduct a pre-experiment to illustrate the motivation behind our framework. As shown in Figure~\ref{fig:fig2}, we evaluate four approaches for leveraging unlabeled data: (1) a baseline excluding unlabeled data, (2) high-confidence pseudo-labeling, (3) domain alignment, and (4) our proposed method. The results reveal that conventional pseudo-labeling often favors easily classified samples, which are typically simplistic pseudo-samples with limited generalization. Additionally, the domain gap causes visual embeddings in $\mathcal{D}_t$ to diverge from those of the source domain's trained model, leading to unreliable feature alignment.

Inspired by this, we propose using text clues as a bridge to coordinate source-target domain knowledge, improve distribution shift, and introduce a curriculum learning strategy to dynamically integrate high-value difficult samples. The DPGNet consists of two stages: source domain pre-training and joint domain adaptation. In the first stage, to establish text-guided domain alignment, given the pre-processed source domain image $x_s \in \mathcal{D}_s$, we use the visual encoder $E_v$ to extract semantically rich features \(z_s^f \in \mathbb{R}^{256 \times 1024}\)  from it, and align $z_s^f$ with the real/fake specific text vectors \(e_t^f \in \mathbb{R}^{768}\) generated by the hint learning module. This alignment is optimized through a composite constraint $\mathcal{L}_{\text{Source}}$, ensuring that $z_s^f$ captures category-independent semantic representations, while $f_r, f_f$ encodes domain-invariant features of real and fake samples.

In the second phase, for unlabeled images from the target domain $x_t \in \mathcal{D}_t$, the DPGNet extracts features using a fine-tuned encoder $E_v$ to obtain $z_t^{f}$, perform classification inference, and pad them according to a high confidence threshold $\lambda_{tf}$, thus obtaining a feature base for the target domain $B_{\text{real}}, B_{\text{fake}}$. To generate reliable pseudo labels, we measure the feature distances between $z_t^{f}$ and $B_{\text{real}}, B_{\text{fake}}$ in the feature library and combine them with the classifier predictions to obtain pseudo labels. To improve the quality of pseudo labels, we introduce a curriculum learning strategy to dynamically adjust the screening threshold $\lambda_{lt}$ to merge challenging high-value samples. To mitigate catastrophic forgetting, we apply cross-domain augmentation to source domain features \(z_s^f\) and employ knowledge distillation to align representations of \(\mathcal{D}_s\) and \(\mathcal{D}_u\), enabling joint training and enhancing generalization across domains.

\subsection{Text-Guided Cross-Domain Alignment}
In this module, the visual encoder is jointly trained with learnable text prompts. Drawing inspiration from prior work~\cite{2025unlocking}, we selectively fine-tune the layer normalization parameters of the first 24 Transformer layers to preserve pre-trained knowledge. For a source domain image \(x_s \in \mathcal{D}_s\), the visual encoder \(E_v\) extracts visual embedding features \(z_s^f \in \mathbb{R}^{256 \times 1024}\), which are processed by a classification head \(h(\cdot)\) to produce the prediction \(\hat{y}_s = h(z_s^f)\). This process enables \(E_v\) to effectively capture forgery-related features, which can be formulated as:
\begin{equation}
\mathcal{L}_{\text{cls}} = \frac{1}{N_s} \sum_{i=1}^{N_s} w_i \cdot \text{CE}(\hat{y}_{s,i}, y_{s,i})
\end{equation}
where \(w_i\) is initially set to 1. To mitigate the imbalance in the sample sources domain and enhance the semantic representation of the real faces, we assign a higher learning weight \(w_i = 2.0\) to the real samples (\(y_{s,i} = 0\)).

\noindent\textbf{Learnable Text Prompts.}
We introduce two trainable text prompts, initialized as "real face photo" and "deep fake face photo", parameterized as \(v_{r} \in \mathbb{R}^{d}\) and \(v_f \in \mathbb{R}^{d}\), where \(d\) is the text embedding dimension. These prompts are fed into the CLIP text encoder \(E_t\) to generate concept embeddings:
\begin{equation}
\textbf{$f_{r} = f_t(v_r), \quad f_{f} = f_t(v_f)$}
\end{equation}
where these embeddings serve as semantic anchors for real and fake categories, \textbf{capturing domain-invariant concepts} that transcend source-target distribution shifts.

\noindent\textbf{Visual-Text Alignment.}
To ensure that the semantics encoded by the visual feature \(z_i^f\) are compatible with {the text embedding $f_{y_i}$}, we perform visual-text alignment:
\label{eq:align}
\begin{equation}
\mathcal{L}_{\text{alig}} = -\log \frac{\exp(\text{sim}(z_i^f, f_{y_i}) / \tau)}{\sum_{c \in \{r, f\}} \exp(\text{sim}(z_i^f, f_c) / \tau)}
\end{equation}
where \(\text{sim}(\cdot, \cdot)\) denotes cosine similarity, and \(f_{y_i}\) is the text embedding corresponding to the ground-truth label \(y_i\). This alignment minimizes domain-specific biases, ensuring that the visual encoder's learned representation \(\mathcal{F}_v = \{z_i^f \mid x_i \in \mathcal{D}_i\}\) is invariant to domain-specific artifacts \(\mathcal{A}_d \subseteq \mathcal{F}_v\) irrelevant to the classification task.

\noindent\textbf{Textual Contrastive Enhancement.}
To enhance the discriminative power of \(\mathcal{F}_v\) and promote class cohesion and separation, we apply a contrastive constraints:
\begin{equation}
\label{eq:con}
\mathcal{L}_{{con}} = \mathbb{E}_{x_i \sim \mathcal{D}_i} \left[ -\text{sim}(z_i^f, f_{y_i}) + \text{sim}(z_i^f, \textbf{$f_{\neg y_i}$}) \right]
\end{equation}
where \textbf{$f_{\neg y_i}$} is the text embedding of the opposite class. This process prioritizes task-relevant semantic features \(\mathcal{S}_c \subseteq \mathcal{F}_v\) (related to authenticity) over domain-specific features \(\mathcal{A}_d\).

\subsection{Curriculum-Driven Pseudo Label Generation}
In this module, we design a curriculum learning strategy with dynamic threshold supervision to address the limitations of common pseudo-label-based strategies.

\noindent\textbf{Feature library construction.} We extract visual features $z_t^f = f_v(x_t)$ for all target samples and generate initial pseudo-labels and their confidence scores for each sample through the classification head \(h(\cdot)\):
\begin{equation}
(\lambda_{\text{real}}^{j}, \lambda_{\text{fake}}^{j}) = h(z_t^f), \quad \lambda_{j} = \max(\lambda_{\text{real}}^{j}, \lambda_{\text{fake}}^{j})
\end{equation}
where we retain samples of $\lambda_{{j}}\geq \lambda_{tf}$ to construct the feature library $\mathcal{B}$. $\mathcal{B}$ is split into real/fake sub-libraries $\mathcal{B}_{\text{real}}$ and $\mathcal{B}_{\text{fake}}$ according to the pseudo labels, where $\mathcal{B}_{\text{real}}$ and $\mathcal{B}_{\text{fake}}$ contain features labeled as real and fake according to the initial visual encoder predictions, respectively, which we use as a ‘simple’ reference case for curriculum learning.

\noindent\textbf{Dynamic Threshold for Pseudo Label Generation.}
For each target sample \(x_t^j\), we generate pseudo labels through a dual-verification process. First, consistent with the calculation in the feature library construction, we get the CLIP-based prediction {$\hat{y}_{{c}}^j = \arg\max(\lambda_{{real}}^j, \lambda_{{fake}}^j)$ and its confidence $\lambda_j = \max(\lambda_{{real}}^j, \lambda_{{fake}}^j)$}. 
Next, we assess the feature library distance by calculating the minimum L2 distance between \(z_t^f\) and the features in fake sub-libraries:
\begin{equation}
d_{f}^j = \min_{\mathbf{z} \in \mathcal{B}_{{fake}}} \|\mathbf{z} - z_t^f\|_2
\end{equation}
where we assign \(\hat{y}_{b}^j = 0\) if \(d_{f}^j > 0.5\), or \(\hat{y}_{b}^j = 1\) otherwise. A pseudo label \(\hat{y}_{unlab}^j\) is accepted if \(\hat{y}_{c}^j = \hat{y}_{b}^j\) and \textbf{the confidence $\lambda_j \geq \lambda_{lt}^{(t)}$}, 
where \(\lambda_{lt}^{(t)}\) is a dynamic threshold. Based on the analysis of simple samples in the feature pool, we initialize \(\lambda_{lt}^{(0)} = 0.85\), which gradually decreases to 0.70 during training. This curriculum strategy initially prioritizes easier samples and progressively incorporates more challenging samples as \(\lambda_{lt}^{(t)}\) decreases, ensuring learning from diverse target features \(\mathcal{F}_t = \{z_t^f \mid x_t \in \mathcal{D}_t\}\) while minimizing bias toward less informative samples.

\subsection{End-to-End Training Strategy}
\noindent\textbf{Latent Space Domain Augmentation.}
Grounded in domain adaptation theory, the transition from source to target domain training often introduces conflicts due to distributional disparities, leading to degraded performance. Therefore, we propose a cross-domain augmentation strategy that integrates latent representations into the source domain’s forgery feature space. By augmenting the source domain's feature space with target domain information, we expand the latent feature space of training samples and create an intermediate representation that bridges the two domains. This facilitates a smoother learning process, avoiding abrupt shifts between domains. Specifically, we compute a linear combination of latent features \(z_s^f\) and \(z_t^f\), extracted from source samples \(x_s \in \mathcal{D}_s\) and target samples \(x_t \in \mathcal{D}_t\):
\begin{equation}
\ z_d^{f} = \alpha z_s^f + (1 - \alpha) z_t^f, \quad i \neq k, \quad \alpha \sim \text{Uniform}(0, 1)
\end{equation}
where \( \alpha \) controls the interpolation weights. Learning the augmented features \(\ z_d^{f}\) strengthens the decision boundary and preserves shared feature structures across domains. We define this process as follows:
\begin{equation}
\mathcal{L}_{\text{dis}} = \mathbb{E}_{x_s \in \mathcal{D}_s, x_t \in \mathcal{D}_t} \left[ \| z_s^{f} - z_d^f \|_2^2 \right]
\end{equation}
\noindent\textbf{End-to-End Loss Design.}
The overall training goal integrates the learning of the source domain and the target domain. The first stage is source domain training, and the second stage is joint training. The whole process is end-to-end, and the specific training objectives are as follows:
\begin{equation}
\mathcal{L}_{\text{total}} = \mathcal{L}_{\text{p1}} + \mathcal{L}_{\text{p2}}
\end{equation}
\begin{equation}
\mathcal{L}_{\text{p1}} = \mathcal{L}_{\text{cls}}^{\text{s}} + \lambda \mathcal{L}_{\text{alig}}^{\text{s}}
\end{equation}
\begin{equation}
\mathcal{L}_{\text{p2}} = \mathcal{L}_{\text{pse}} + \lambda_1 \mathcal{L}_{\text{cls}}^{\text{s}} + \lambda_2 \mathcal{L}_{\text{alig}}^{\text{s}} + \beta \mathcal{L}_{\text{dis}}
\end{equation}
\begin{equation}
\mathcal{L}_{\text{pse}} = \mathcal{L}_{\text{con}}^{\text{t}} + \mathcal{L}_{\text{cls}}^{\text{t}} + \mathcal{L}_{\text{alig}}^{\text{t}}
\end{equation}
where \(\lambda\), \(\lambda_1\), \(\lambda_2\) and \(\beta\) are weight factors for balancing. The specific settings are detailed in the experimental section.

%% file: 4_experiments.tex
\section{Experiments}
\subsection{Settings}
\noindent\textbf{Datasets.} We used the following nine datasets: FaceForensics++(FF++) \cite{ff++}, Deepfake Detection Challenge(DFDC) \cite{dfdc}, preview version of DFDC(DFDCP) \cite{dfdcp}, two versions of CelebDF (CDF-v1, CDF-v2) \cite{li2019celeb,li2020celeb}, DeepfakeDetection (DFD) \cite{dfd}, DF40 \cite{yan2024df40}, UCDDP and UDF40, respectively; UCDDP is an unlabeled dataset obtained by sampling a small amount from the training sets of Deepfake Detection Challenge(DFDC), preview version of DFDC (DFDCP), two versions of CelebDF (CDF-v1, CDF-v2), DeepfakeDetection (DFD), and UDF40 is an unlabeled dataset obtained by sampling a small amount from the data subsets of different counterfeiting methods of DF40.

\begin{table}[!t]
% \vspace{-0.4in}
\centering
\resizebox{\linewidth}{!}{
\begin{tabular}{@{}ccccccc@{}}  % 列数保持7列（新增比例列）
    \toprule
    \multirow{2}{*}{\textbf{Dataset}} & \multirow{2}{*}{\textbf{\begin{tabular}[c]{@{}c@{}}Real\\ Videos\end{tabular}}} & \multirow{2}{*}{\textbf{\begin{tabular}[c]{@{}c@{}}Fake\\ Videos\end{tabular}}} & \multirow{2}{*}{\textbf{\begin{tabular}[c]{@{}c@{}}Total\\ Videos\end{tabular}}} & \multirow{2}{*}{\textbf{\begin{tabular}[c]{@{}c@{}}Synthesis\\ Methods\end{tabular}}} & \multirow{2}{*}{\textbf{\begin{tabular}[c]{@{}c@{}}Real:Fake\\ Ratio\end{tabular}}} & \multirow{2}{*}{\textbf{\begin{tabular}[c]{@{}c@{}}Total\\ Image \end{tabular}}} \\ 
    \\ \midrule
    {FF++~\cite{rossler2019faceforensics++}}     & 1000         & 4000      & 5000     & 4    & 1:4     & 320k     \\
    {CelebDF-v1~\cite{li2020celeb}} $\dagger$       & 408         & 795      & 1203    & 1    & $\sim$1:2       & 77k     \\
    {CelebDF-v2~\cite{li2020celeb}} $\dagger$       & 590        & 5639      & 6229     & 1    & $\sim$1:10      & 399k     \\
    {DFDCP~\cite{dfdcp}} $\dagger$    & 1131         & 4119      & 5250   & 2    & $\sim$1:4     & 336k     \\
    {DFDC~\cite{dolhansky2020deepfake}} $\dagger$    & 23654        & 104500     & 128154    & 8    & $\sim$1:5  & 8202k     \\
    {DFD~\cite{dfd}} $\dagger$      & 363       & 3000    & 3363     & 5    & $\sim$1:10      & 215k     \\
    {DF40~\cite{yan2024df40}} $*$       & $\sim$1500        & 0.1M+       & 0.1M+     & 40   & $\sim$1:40 & 1M+     \\
    \midrule
    {UCDDP} $\dagger$     & --         & --      & --     & 10+    & $\sim$1:5           & 18k    \\
    {UDF40} $*$      & --         & --    & --    & 6    & $\sim$1:8           & 9k     \\ \bottomrule
 \end{tabular}}
\caption{Details of the dataset used. UCDDP and UDF40 are uniformly sampled from the above datasets. The symbols $*$ and $\dagger$ represent the corresponding sampling relationship.}
\label{tab:dataset_comparison}
\vspace{-0.2in}
\end{table}

\begin{table*}[!t]
\vspace{-0.2in}
\centering
\scalebox{1.0}{
\begin{tabular}{l|c|c|c|c|c|c|c|c}
\toprule
\textbf{Methods} & \textbf{Venue} & \textbf{Backbone} & \textbf{CDF-v1} & \textbf{CDF-v2} & \textbf{DFD} & \textbf{DFDC} & \textbf{DFDCP} & \textbf{Avg.       } \\
\midrule
FFD~\cite{ffd} &CVPR'20 & Xception & 0.784 & 0.744 & 0.802 & 0.703 & 0.743 & 0.755 \\
SRM~\cite{luo2021generalizing} &CVPR'21  & Xception & 0.793 & 0.755 & 0.812 & 0.700 & 0.741 & 0.760 \\
SPSL~\cite{spsl} &CVPR'21 & Xception & 0.815 & 0.765 & 0.812 & 0.704 & 0.741 & 0.767 \\
RECCE~\cite{cao2022end} & CVPR'22 & Designed & 0.768 & 0.732 & 0.812 & 0.713 & 0.734 & 0.752 \\
CORE~\cite{ni2022core} & CVPR'22  & Xception & 0.780 & 0.743 & 0.802 & 0.705 & 0.734 & 0.753 \\
UCF~\cite{yan2023ucf} & ICCV'23 & Xception & 0.779 & 0.753 & 0.807 & 0.719 & 0.759 & 0.763 \\
ED$\dagger$~\cite{ba}&AAAI'24 &ResNet-34 & 0.818&{0.864} & - & 0.721 & {0.851} & -  \\ %原论文
LSDA~\cite{lsda} &CVPR'24 &EfficientNet-B4 &0.867& 0.830 & 0.880 & 0.736  &0.815  & 0.826 \\%原论文
ProDet$\dagger$~\cite{cheng2024can}& NeurIPS'24 &EfficientNet-B4 & 0.909 &0.844& -  & 0.811  & 0.724 &0.822 \\ %o 原论文
UDD$\dagger$~\cite{udd}&AAAI'25  &ViT-B/16 &- & 0.869 & 0.910 & 0.758  &0.856 & - \\ %原论文
Effort~\cite{effort}&ICML'25 & CLIP (ViT-L/14)  &0.926 &0.878 & 0.922 & 0.822 & 0.835  &0.877 \\ %原论文
F-Adapter~\cite{ForensicsAdapter}&CVPR'25 & CLIP (ViT-L/14)  &0.914 &0.900 & 0.933 & 0.843 & 0.890  &0.896 \\ %原论文

\midrule
\multirow{2}*{DPGNet (\textbf{ours})} & \multirow{2}*{-} & \multirow{2}*{CLIP (ViT-L/14)} & \textbf{0.973} & \textbf{0.957} & \textbf{0.951} & \textbf{0.892} & \textbf{0.917} & \textbf{0.938} \\
& & & \textbf{\textcolor{cvprgreen}{(↑4.7\%)}} & \textbf{\textcolor{cvprgreen}{(↑5.7\%)}} & \textbf{\textcolor{cvprgreen}{(↑1.8\%)}} & \textbf{\textcolor{cvprgreen}{(↑4.9\%)}} & \textbf{\textcolor{cvprgreen}{(↑2.7\%)}} & \textbf{\textcolor{cvprgreen}{(↑4.2\%)}} \\
\bottomrule
\end{tabular}  
}
 \vspace{-2mm}
\caption{Benchmark results for cross-dataset evaluation (Protocol‑1, frame-level AUC). All detectors are trained on FF++ c23~\cite{ff++} and evaluated on other deepfake datasets. $\dagger$ indicates that the result is taken from the original paper.}
\label{tab:cross-dataset}
\end{table*}

\begin{table*}[!t]
\vspace{-3mm}
\centering
\scalebox{0.95}{
\begin{tabular}{l|c|c|c|c|c|c|c|c}
\toprule
\textbf{Methods} & \textbf{Backbone} & \textbf{UniFace} & \textbf{BleFace} & \textbf{MobSwap} & \textbf{FaceDan} & \textbf{InSwap} & \textbf{SimSwap} & \textbf{Avg.} \\
\midrule
SPSL~\cite{liu2021spatial} & Xception & 0.747 & 0.748 & 0.885 & 0.666 & 0.643 & 0.665 & 0.726 \\
SRM~\cite{luo2021generalizing} & Xception & 0.749 & 0.704 & 0.779 & 0.659 & 0.793 & 0.694 & 0.730 \\
CORE~\cite{ni2022core} & Xception & 0.871 & 0.843 & 0.959 & 0.774 & 0.855 & 0.724 & 0.838 \\
RECCE~\cite{cao2022end} & Designed & 0.898 & 0.832 & 0.925 & 0.848 & 0.848 & 0.768 & 0.853 \\
SLADD~\cite{chen2022self} &Xception  & 0.878 & 0.882 & 0.954 & 0.825 & 0.879 & 0.794 & 0.869 \\
SBI~\cite{shiohara2022detecting} &EfficientNet-B4  & 0.724 & 0.891 & 0.952 & 0.594 & 0.712 & 0.701 & 0.762 \\
UCF~\cite{yan2023ucf} & Xception & 0.831 & 0.827 & 0.950 & 0.862 & 0.809 & 0.647 & 0.821 \\
IID~\cite{huang2023implicit} &Designed  & 0.839 & 0.789 & 0.888 & 0.844 & 0.789 & 0.644 & 0.799 \\
LSDA~\cite{lsda} & EfficientNet-B4 & 0.872 & 0.875 & 0.930 & 0.721 & 0.855 & 0.793 & 0.841 \\
ProDet~\cite{cheng2024can} & EfficientNet-B4 & 0.908 & 0.929 & 0.975 & 0.747 & 0.837 & 0.844 & 0.873 \\
CDFA~\cite{lin2024fake} & SwinV2-B   & 0.762 & 0.756 & 0.823 & 0.803 & 0.772 & 0.757 & 0.779 \\
F-Adapter~\cite{ForensicsAdapter} & CLIP (ViT-L/14) & 0.969 & 0.886 & 0.963 & 0.943 & 0.937 & 0.917 & 0.936 \\
Effort~\cite{effort} & CLIP (ViT-L/14) & 0.962 & 0.873 & 0.953 & 0.926 & 0.936 & 0.926 & 0.929 \\
\midrule
\multirow{2}*{DPGNet (\textbf{ours})} & \multirow{2}*{CLIP (ViT-L/14)} & \textbf{0.987} & \textbf{0.984} & \textbf{0.990} & \textbf{0.974} & \textbf{0.972} & \textbf{0.984} & \textbf{0.982} \\
& & \textbf{\textcolor{cvprgreen}{(↑1.86\%)}} & \textbf{\textcolor{cvprgreen}{(↑5.92\%)}} & \textbf{\textcolor{cvprgreen}{(↑1.54\%)}} & \textbf{\textcolor{cvprgreen}{(↑3.29\%)}} & \textbf{\textcolor{cvprgreen}{(↑3.74\%)}} & \textbf{\textcolor{cvprgreen}{(↑6.26\%)}} & \textbf{\textcolor{cvprgreen}{(↑4.91\%)}} \\
\bottomrule
\end{tabular}
}
\caption{Benchmarking Results for Cross-Method Evaluation (Protocol-2, Video-Level AUC). All detectors are trained on FF++ c23~\cite{rossler2019faceforensics++} and evaluated on other deepfake datasets.}
\label{tab:cross-method}
  \vspace{-3mm}
\end{table*}

\noindent\textbf{Evaluation Protocol.}The results in Figure~\ref{fig:fig2} show that simply using high-confidence pseudo labels to align with the domain does not lead to improved performance, especially for detectors with lower performance, a large number of pseudo-label generation errors can even lead to performance degradation. Accordingly, we adopt two widely used standard protocols for evaluation: Protocol 1 is used for cross-dataset evaluation, and Protocol 2 is used for cross-operation evaluation within the FF++ domain. For Protocol 1, the model is trained using the labeled source domain dataset FF++ and the small unlabeled dataset UCDDP, and the performance is evaluated on the test set corresponding to the UCDDP sampling dataset (DFDC, DFDCP, CDF-v1, CDF-v2, DFD) to evaluate the generalization ability across datasets. For Protocol 2, the model is trained using FF++ and the small unlabeled dataset UDF40, and the performance is evaluated on the test subset corresponding to the UDF40 sampling dataset DF40 to evaluate the generalization ability across different forgery methods under a consistent data distribution. \textbf{To further demonstrate that our performance improvement is not due to the use of unlabeled data}, we let the baseline methods use the same unlabeled dataset (UCDDP or UDF40) for additional comparison by generating and utilizing pseudo labels.

\noindent\textbf{Implementation Details.}
We adopt CLIP ViT-L/14~\cite{clip_paper} as the visual backbone, with input images resized to 224$\times$224 pixels. During training, we sample 16 frames per video, while 32 frames are used for testing. The model is optimized using the Adam optimizer~\cite{adam} with a learning rate of 0.00008 and a weight decay of 0.0005. For training, the batch size is set to 32 for the source domain (FF++) and 10 for unlabeled data (UCDDP or UD40), with a test batch size of 32. Standard data augmentation techniques, including random cropping and flipping, are applied to enhance data diversity. For feature library construction, we set the initial confidence threshold to \(\lambda_{tf} = 0.9\). The dynamic pseudo-labeling threshold \(\lambda_{\text{lt}}\) starts at 0.85 and gradually decreases to 0.70 during training. Loss hyperparameters \(\lambda\), \(\lambda_1\), \(\lambda_2\), and \(\beta\) are empirically set to 0.8, 0.4, 0.5, and 0.1, respectively. For evaluation, we report frame-level and video-level Area Under the Curve (AUC), a standard metric in deepfake detection, to compare our method with prior work. AUC provides a robust measure of classification performance across varying thresholds. All experiments are conducted on an NVIDIA RTX 4090 GPU.
\begin{table}
    \centering
    \setlength{\tabcolsep}{1pt} % 减小列间距以缩短宽度
    \scalebox{0.65}{ % 略微放大以适应内容
        \begin{tabular}{c|c|c|ccccccc}
            \toprule
            \multirow{2}{*}{{Methods}} & \multirow{2}{*}{{Train Set}} & \multirow{2}{*}{{\(\lambda\)}} & \multicolumn{7}{c}{{Cross-method Evaluation}} \\
            \cmidrule(lr){4-10}
            & & & UniFace & BleFace & MobSwap & FaceDan & InSwap & SimSwap & {Avg.} \\
            \midrule
            \multirow{4}{*}{\begin{tabular}[c]{@{}c@{}}{F-Ada}\\\end{tabular}} 
            & FF++ &- & 0.919 & 0.818 & 0.940 & 0.904 & 0.904 & 0.856 & 0.890 \\
            & +UDF40 & 0.9 & 0.924 & 0.837 & 0.946 & 0.912 & 0.916 & 0.876 & 0.902 \\
            & +UDF40 & 0.8 & 0.929 & 0.837 & 0.947 & 0.917 & 0.919 & 0.862 & 0.902 \\
            & +UDF40 & 0.7 & 0.924 & 0.835 & 0.947 & 0.912 & 0.921 & 0.867 & 0.901 \\
            \midrule
            \multirow{4}{*}{\begin{tabular}[c]{@{}c@{}}{Effort}\\\end{tabular}} 
            & FF++ &- & 0.940 & 0.825 & 0.911 & 0.883 & 0.907 & 0.885 & 0.892 \\
            & +UDF40 & 0.9 & 0.932 & 0.852 & 0.918 & 0.897 & 0.899 & 0.889 & 0.898 \\
            & +UDF40 & 0.8 & 0.938 & 0.837 & 0.928 & 0.896 & 0.908 & 0.899 & 0.901 \\
            & +UDF40 & 0.7 & 0.940 & 0.841 & 0.932 & 0.897 & 0.910 & 0.901 & 0.904 \\
            \midrule
            \multirow{2}{*}{{Ours}} & \multirow{2}{*}{+UDF40} & \multirow{2}{*}{Adp} & \textbf{0.972} & \textbf{0.971} & \textbf{0.981} & \textbf{0.954} & \textbf{0.952} & \textbf{0.974} & \textbf{0.967} \\
            & & & \textbf{\textcolor{darkred}{(↑3.2\%)}} & \textbf{\textcolor{darkred}{(↑11.9\%)}} & \textbf{\textcolor{darkred}{(↑4.9\%)}} & \textbf{\textcolor{darkred}{(↑3.7\%)}} & \textbf{\textcolor{darkred}{(↑4.2\%)}} & \textbf{\textcolor{darkred}{(↑7.3\%)}} & \textbf{\textcolor{darkred}{(↑6.3\%)}} \\
            \bottomrule
        \end{tabular}
    }
    \caption{Cross-dataset evaluation of the baseline methods using unlabeled data (Frames-level AUC). U represents the unlabeled sampling data set corresponding to the test set, and \(\lambda\) denotes the confidence threshold used in training.}
    \label{tab:table3}
\vspace{-0.3cm}
\end{table}

\begin{table}
% \vspace{-0.4in}
    \centering
    \setlength{\tabcolsep}{1pt} %
    \scalebox{0.75}{ 
        \begin{tabular}{c|c|c|cccccc}
            \toprule
            \multirow{2}{*}{{Methods}} & \multirow{2}{*}{{Train Set}} & \multirow{2}{*}{{\(\lambda\)}} & \multicolumn{6}{c}{{Cross-dataset Evaluation}} \\
            \cmidrule(lr){4-9}
            & & & CDF-v1 & CDF-v2 & DFD & DFDC & DFDCP & {Avg.} \\
            \midrule
            \multirow{4}{*}{\begin{tabular}[c]{@{}c@{}}{F-Ada}\\\end{tabular}} 
            & FF++ & - & 0.914 & 0.900 & 0.933 & 0.843 & 0.890 & 0.896 \\
            & +UCDDP & 0.9 & 0.903 & 0.905 & 0.924 & 0.835 & 0.874 & 0.888 \\
            & +UCDDP & 0.8 & 0.936 & 0.906 & 0.927 & 0.834 & 0.868 & 0.894 \\
            & +UCDDP & 0.7 & 0.924 & 0.897 & 0.867 & 0.842 & 0.885 & 0.883 \\
            \midrule
            \multirow{4}{*}{\begin{tabular}[c]{@{}c@{}}{Effort}\\\end{tabular}} 
            & FF++ & - & 0.926 & 0.872 & 0.922 & 0.822 & 0.835 & 0.875 \\
            & +UCDDP & 0.9 & 0.924 & 0.907 & 0.929 & 0.833 & 0.845 & 0.888 \\
            & +UCDDP & 0.8 & 0.935 & 0.901 & 0.920 & 0.840 & 0.842 & 0.888 \\
            & +UCDDP & 0.7 & 0.933 & 0.891 & 0.912 & 0.829 & 0.830 & 0.879 \\
            \midrule
            \multirow{2}{*}{{Ours}} & \multirow{2}{*}{+UCDDP} & \multirow{2}{*}{Adp} & \textbf{0.973} & \textbf{0.957} & \textbf{0.951} & \textbf{0.892} & \textbf{0.917} & \textbf{0.938} \\
            & & & \textbf{\textcolor{darkred}{(↑3.7\%)}} & \textbf{\textcolor{darkred}{(↑5.0\%)}} & \textbf{\textcolor{darkred}{(↑2.2\%)}} & \textbf{\textcolor{darkred}{(↑5.2\%)}} & \textbf{\textcolor{darkred}{(↑2.7\%)}} & \textbf{\textcolor{darkred}{(↑4.2\%)}} \\
            \bottomrule
        \end{tabular}
    }
    \caption{Cross-method evaluation of baseline methods using unlabeled data (Frames-level AUC). U represents the unlabeled sampling data set corresponding to the test set, and \(\lambda\) denotes the confidence threshold used in training.}
    \label{tab:table4}
\vspace{-0.3cm}
\end{table}

\begin{table*}[!t]
\vspace{-0.25in}
\setlength{\tabcolsep}{4pt} 
\small%
\centering
\begin{tabular}{c|ccc|ccc|ccc|ccc|ccc}
\toprule
 \multirow{2}{*}{Number of samples}  & \multicolumn{3}{c|}{CDF-v1} & \multicolumn{3}{c|}{CDF-v2} & \multicolumn{3}{c|}{DFDC} & \multicolumn{3}{c|}{DFDCP} & \multicolumn{3}{c}{DFD} \\ 
\cmidrule(lr){2-4} \cmidrule(lr){5-7} \cmidrule(lr){8-10} \cmidrule(lr){11-13} \cmidrule(lr){14-16} 
 & AUC & AP & EER & AUC & AP & EER & AUC & AP & EER & AUC & AP & EER & AUC & AP & EER  \\
\midrule
 $6k$  & 0.975 & 0.984 & 8.0 & 0.940 & 0.968 & 13.4 & 0.862 & 0.889 & 22.3 & 0.876 & 0.936 & 20.6 & 0.947 & 0.993 & 11.1 \\  
$12k$ & \textbf{0.985} & \textbf{0.990} & \textbf{6.6} & 0.952 & 0.975 & 11.5 & 0.867 & 0.891 & 21.9 & 0.891 & 0.943 & 19.5 & 0.939 & 0.992 & 12.7 \\ 
$18k$ & 0.973 & 0.985 & 8.7 & 0.957 & 0.978 & 11.0 & \textbf{0.892} & \textbf{0.914} & \textbf{19.1} & \textbf{0.917} & \textbf{0.956} & \textbf{16.8} & \textbf{0.951} & \textbf{0.994} & \textbf{10.4} \\
$24k$ & 0.975 & 0.984 & 8.6 & \textbf{0.964} & \textbf{0.979} & \textbf{9.6} & 0.883 & 0.906 & 20.0 & 0.910 & 0.952 & 17.0 & 0.946 & 0.993 & 11.1 \\ %avg auc 0.896 ap 0.938 eer18.2 
\bottomrule
\end{tabular}
\vspace{-0.1in}
\caption{Ablation study on the number of samples of unlabeled datasets, evaluated using frame-level AUC, AP, and EER.}
\label{tab: sample number1}
% \vspace{-0.2in}
\end{table*}

\begin{table*}[!t]
\vspace{-0.1in}
\setlength{\tabcolsep}{4pt} % 列间距
\small%
\centering
  \scalebox{0.85}{
\begin{tabular}{c|ccc|ccc|ccc|ccc|ccc|ccc}
\toprule
 \multirow{2}{*}{Number of samples}   & \multicolumn{3}{c|}{UniFace} & \multicolumn{3}{c|}{BleFace} & \multicolumn{3}{c|}{MobSwap} & \multicolumn{3}{c|}{FaceDan} & \multicolumn{3}{c}{InSwap}& \multicolumn{3}{|c}{SimSwap} \\ 
\cmidrule(lr){2-4} \cmidrule(lr){5-7} \cmidrule(lr){8-10} \cmidrule(lr){11-13} \cmidrule(lr){14-16} \cmidrule(lr){17-19} 
 & AUC & AP & EER & AUC & AP & EER & AUC & AP & EER & AUC & AP & EER & AUC & AP & EER  & AUC & AP & EER \\
\midrule
 $4k$  & 0.959 & 0.965 & 8.4 & 0.938 & 0.944 & 12.1 & 0.973 & 0.995 & 7.0 & 0.928 & 0.934 & 12.9& 0.947 &  0.942 & 10.2 & 0.955 &  0.960 & 9.1\\  

$8k$  & 0.970 &0.971 & \textbf{7.3} & 0.964 &  0.963 & 8.7 &0.976& 0.995 & 6.9 & 0.938 &0.940& 11.3  & 0.949 &0.942& 10.3& 0.968 &  0.967 & 7.6 \\ %avg auc 0.832 ap 0.888 eer 24.1

$12k$  & 0.972 &0.971 &7.7 &0.968 & 0.970 & 8.2 & 0.978 & 0.996& 6.6 & 0.954 &0.958 & 10.1 &  0.953 &0.950 &10.1& 0.973 &  0.975 & 6.9\\
$16k$ &  \textbf{0.973} & \textbf{0.973} & 7.9 & \textbf{0.971} & \textbf{0.972} & \textbf{7.3} & \textbf{0.981} & \textbf{0.997} & \textbf{6.4} & \textbf{0.963} & \textbf{0.964} & \textbf{8.8} & \textbf{0.957} & \textbf{0.954}& \textbf{9.1} & \textbf{0.976} &  \textbf{0.994} & \textbf{6.5}\\ %avg auc 0.896 ap 0.938 eer18.2 
\bottomrule
\end{tabular}
}
\vspace{-0.1in}
\caption{Ablation study on the number of samples of unlabeled datasets, evaluated using frame-level AUC, AP, and EER.}
\label{tab:sample number2}
\vspace{-0.3cm}
\end{table*}

\begin{table}[!t]
  \centering
  \setlength{\tabcolsep}{9pt}
  \scalebox{0.9}{
  \begin{tabular}{c|c|c|c|c|c} 
  \toprule
    \multicolumn{3}{c|}{Ours} & \multirow{2}{*}{Cross} & \multirow{2}{*}{Cross} & \multirow{2}{*}{Avg.} \\
    \cmidrule(r){1-3}
     \textit{TCA}  & \textit{CPG} & \textit{CD} &-dataset &-method &  \\ 
    \midrule
     $\times$ & $\times$ & $\times$  & 0.871 & 0.857 & 0.864 \\
     $\times$ & \checkmark & \checkmark  & 0.896 & 0.927 & 0.912 \\
     $\times$ & \checkmark & $\times$   & 0.903 & 0.924 & 0.914 \\
     \checkmark & $\times$ & $\times$ & 0.926 & 0.950 & 0.938 \\
     \checkmark & $\times$ & \checkmark  & 0.924 & 0.956 & 0.940 \\
     \checkmark & \checkmark & $\times$  & 0.934 & 0.958 & 0.946 \\
     \checkmark & \checkmark & \checkmark  & 0.938 & 0.967 & 0.953 \\
    \bottomrule
  \end{tabular}
  }
  \vspace{-0.1in}
  \caption{Ablation study on core components. Results for Cross-dataset (UCDDP) and Cross-method (UDF40).
  % Some of the results are cited from \cite{xu2023tall}.
  }
  % \vspace{-0.1in}
  \label{tab:components}
  \vspace{-3mm}
\end{table}

\begin{table}[!t]
  \vspace{-0.05in}
  \centering
  \setlength{\tabcolsep}{9pt}
  \scalebox{0.9}{
  \begin{tabular}{c|c|c|c|c|c} 
  \toprule
    \multicolumn{3}{c|}{Ours} & \multirow{2}{*}{Cross} & \multirow{2}{*}{Cross} & \multirow{2}{*}{Avg.} \\
    \cmidrule(r){1-3}
     $\mathcal{L}_{\rm alig}$  & $\mathcal{L}_{\rm con}$ & $\mathcal{L}_{\rm dis}$ &-dataset &-method &  \\ 
    \midrule
     $\times$ & $\times$ & $\times$  & 0.876 & 0.857 & 0.867 \\
     \checkmark & $\times$& $\times$ & 0.919 & 0.946 & 0.933 \\
     \checkmark & $\times$& \checkmark  & 0.924 & 0.946 & 0.935 \\
     \checkmark & \checkmark& $\times$  & 0.934 & 0.952 & 0.943 \\
     \checkmark & \checkmark& \checkmark  & 0.938 & 0.967 & 0.953 \\
    \bottomrule
  \end{tabular}
  }
  \vspace{-0.1in}
  \caption{Ablation study on embedding alignment ($\mathcal{L}{\rm alig}$), contrast enhancement ($\mathcal{L}{\rm con}$), and cross-domain distillation ($\mathcal{L}_{\rm dis}$). Results for cross-dataset (UCDDP) and cross-method (UDF40).
  % Some of the results are cited from \cite{xu2023tall}.
  }
 \label{tab:loss}
  \vspace{-0.15in}
\end{table}

\subsection{Detection Performance}
% \paragraph{Evaluation Benchmarkin.} We compare with previous competitive detectors, and the results in Table~\ref{tab:cross-dataset} and Table~\ref{tab:cross-method} show the superior performance of our framework. 
Table~\ref{tab:cross-dataset} presents the results of cross-dataset evaluation under Protocol-1. DPGNet achieves an average frame-level AUC of 0.938, surpassing the best baseline, ForensicsAdapter (0.896), by 4.2\%. Notably, DPGNet excels on challenging datasets such as DFDC (AUC of 0.892, +4.9\% over ForensicsAdapter) and CDF-v2 (AUC of 0.957, +5.7\%). This superior performance is attributed to DPGNet’s text-guided alignment technique. For Protocol-2, Table~\ref{tab:cross-method} reports cross-method evaluation results on DF40, where DPGNet achieves an average video-level AUC of 0.982, outperforming ForensicsAdapter (0.936) by 4.91\%. DPGNet demonstrates significant improvements on advanced forgery methods, such as SimSwap (+6.26\%) and BleFace (+5.92\%), highlighting its generalization to diverse manipulation techniques within a consistent domain.

To assess the impact of leveraging unlabeled data, we compare DPGNet against baselines augmented with pseudo-labeling at fixed confidence thresholds (Tables~\ref{tab:table3} and \ref{tab:table4}). For cross-dataset evaluation, ForensicsAdapter with UCDDP (0.7 threshold) suffers a performance drop (average AUC of 0.883, -1.3\% compared to FF++ alone), likely due to noisy pseudo labels. In contrast, DPGNet with UCDDP achieves a robust AUC of 0.938, demonstrating its ability to prioritize high-value samples through dynamic curriculum learning. Similarly, in cross-method evaluation, DPGNet with UDF40 achieves an average AUC of 0.967, surpassing ForensicsAdapter (0.902, +6.5\%) and Effort (0.904, +6.3\%). We may notice that the baseline performance of UCDDP and UDF40 has \textbf{limited improvement and minimal variation} across fixed thresholds, which stems from the limited sample size of these datasets. After threshold-based filtering, the number of usable training samples is further reduced, limiting the impact on models initialized with pre-trained weights, which are inherently stable to small data increments.

% \begin{figure}
% \vspace{-0.05in}
% \centering
% \includegraphics[width=1.0\columnwidth]{sec/radar.png}
% \vspace{-0.2in}
% \caption{Cross-dataset (left) and cross-method evaluation.}
% \vspace{-0.2in}
% \label{fig-leida}
% \end{figure}
\subsection{Ablation Study}
To dissect the contributions of the DPGNet design, we performed ablation studies on the unlabeled sample size, key components, loss functions, and unsupervised baselines.

\noindent\textbf{Unlabeled Sample Size.}
Tables~\ref{tab:table3} and \ref{tab:table4} evaluate the effect of varying unlabeled sample sizes from UCDDP and UDF40. In cross-dataset evaluation, increasing UCDDP samples from 6k to 24k improves the average AUC from 0.896 to 0.938, stabilizing at 18k samples (e.g., DFDCP: AUC 0.914, EER 16.9). In cross-method evaluation, scaling UDF40 samples from 4k to 16k raises the average AUC from 0.950 to 0.970 (e.g., MobSwap: AUC 0.981, EER 6.4). These results demonstrate that our method achieves significant performance gains in target domains with minimal unlabeled data, particularly in cross-method detection, where small sample sizes yield substantial improvements, highlighting the sample efficiency of our domain adaptation approach for deep face forgery detection.

\noindent\textbf{Core Components.}
Table~\ref{tab:components} evaluates the core components of DPGNet: text-guided cross-domain alignment (TCA) and curriculum-driven pseudo label generation (CPG). Additionally, we provide a more in-depth evaluation of the cross-domain distillation (CD) strategy. The baseline without these components yields an AUC of 0.867. TGA alone boosts performance to 0.938  by aligning visual-textual embeddings, mitigating domain gaps. Adding CPL increases the AUC to 0.946 by progressively lowering the pseudo-labeling threshold to include challenging samples, while CD ensures generalization, maintaining the AUC at 0.953. This synergy drives its superior generalization.

\noindent\textbf{Ablation on Loss Functions.}
Table~\ref{tab:loss} evaluates the loss components of DPGNet: embedding alignment (\(\mathcal{L}_{\text{alig}}\)), contrastive enhancement (\(\mathcal{L}_{\text{con}}\)) and distillation across domain (\(\mathcal{L}_{\text{dis}}\)). The baseline without these losses achieves an AUC of 0.867. Adding \(\mathcal{L}_{\text{alig}}\) improves the AUC to 0.933 (+6.6\%) by ensuring domain-invariant representations. Including \(\mathcal{L}_{\text{con}}\) and \(\mathcal{L}_{\text{dis}}\) further stabilizes the performance to 0.953.

\noindent\textbf{Comparison with Unsupervised Methods.}
Table~\ref{tab: unsupervised methods} compares DPGNet against unsupervised methods: DANN \cite{ganin2015unsupervised}, NAMC~\cite{unsurpervised}, and Source-free Domain Adaptation~\cite{li2024comprehensive}. DPGNet outperforms these methods, utilizing text-guided alignment and curriculum learning to capture diverse forgery patterns while preserving source knowledge. 
\begin{table}
  \centering
  \setlength{\tabcolsep}{9pt}
  \scalebox{0.9}{
  \begin{tabular}{c|c|c|c} 
  \toprule
    \multicolumn{1}{c|}{Methods} & \multirow{1}{*}{Cross-dataset} & \multirow{1}{*}{Cross-method} & \multirow{1}{*}{Avg.} \\
    \midrule
     Ori  &
     0.872 & 0.857& 0.865 \\
     DANN\cite{ganin2015unsupervised}  &
     0.846 & 0.838& 0.842 \\

     NAMC~\cite{unsurpervised}  &
     0.878 & 0.877 & 0.878 \\

     SDAT~\cite{li2024comprehensive}   &
     0.864 & 0.842 & 0.853 \\
     
     Ours  &
     0.938 & 0.967 & 0.953 \\
    \bottomrule
  \end{tabular}
  }
\caption{Comparison with domain adaptation methods.} 
  \vspace{-3mm}
  \label{tab: unsupervised methods}
\end{table}

\begin{figure}
\centerline{\includegraphics[width=0.5\textwidth]{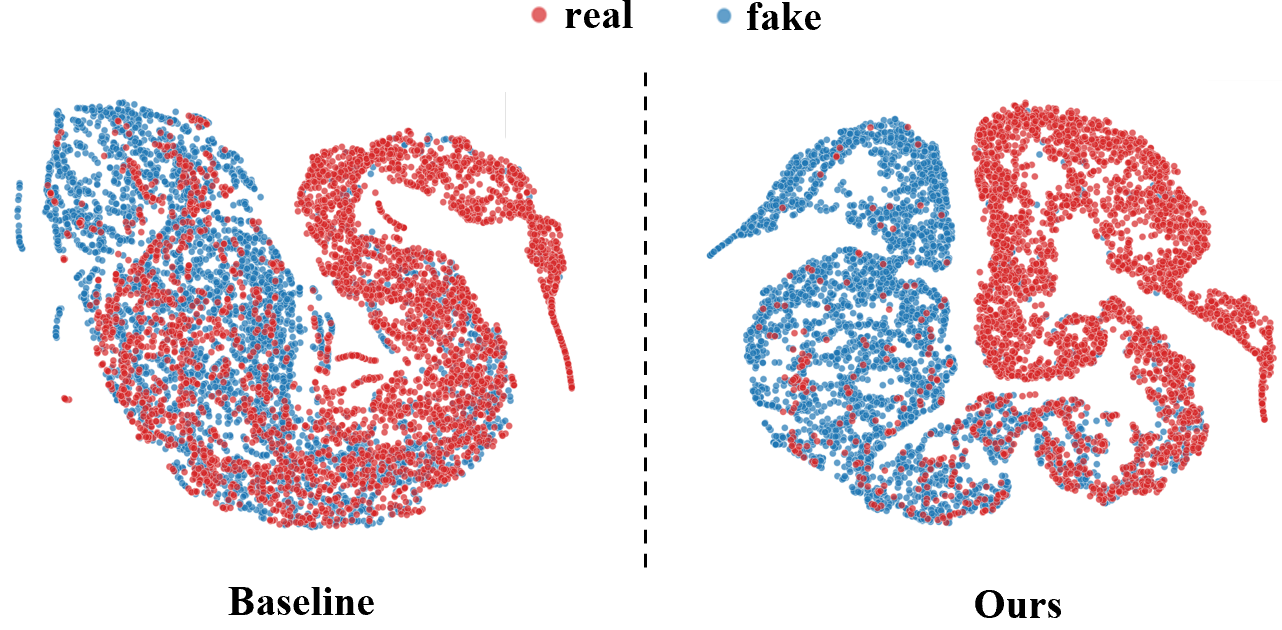}}
\caption{T-SNE visualization on the cross-method test.}
\label{fig:tsne}
  \vspace{-3mm}
\end{figure}

% \begin{table}[!t]
%   \centering
%   \setlength{\tabcolsep}{9pt}
%   \label{tab:ortho_loss}
%   \scalebox{0.75}{
%   \begin{tabular}{c|ccc|ccc}
%     \toprule
%     \multirow{2}{*}{\textbf{\begin{tabular}[c]{@{}c@{}}Screening\\threshold\end{tabular}}} & \multicolumn{3}{c|}{Crosss-Dataset} & \multicolumn{3}{c}{Cross-Method} \\
%     \cmidrule(lr){2-4} \cmidrule(lr){5-7}
%      & AUC & AP & EER & AUC & AP & EER  \\
%     \midrule
%     0.9 & 0.814 & 0.895 & 24.9 & 0.789 & 0.880 & 28.7 \\
%     0.8 & 0.845 & 0.912 & 24.0 & 0.822 & 0.899 & 25.8 \\
%     0.7 & 0.881 & 0.926 & 20.3 & 0.875 & 0.923 & 21.4 \\
%     \scalebox{1.2}{$\lambda_{tf}$}(0.9→0.7) & \textbf{0.914} & \textbf{0.940} & \textbf{15.8} & \textbf{0.900} & \textbf{0.945} & \textbf{18.2} \\
%     \scalebox{1.2}{$\lambda_{tf}$}(0.9→0.65) & \textbf{0.914} & \textbf{0.940} & \textbf{15.8} & \textbf{0.900} & \textbf{0.945} & \textbf{18.2} \\
%     \scalebox{1.2}{$\lambda_{tf}$}(0.9→0.6) & \textbf{0.914} & \textbf{0.940} & \textbf{15.8} & \textbf{0.900} & \textbf{0.945} & \textbf{18.2} \\
%     \bottomrule
%   \end{tabular}
%   }
%   \caption{\textbf{Ablation study of pseudo-label threshold. CDF-v2 and SimSwap are the test results after using UCDDP and UDF40 respectively.}}
%   \vspace{-3mm}
% \end{table}

\subsection{Feature Distribution Visualization}
To show the uniqueness of DPGNet from the feature distribution level, we use T-SNE \cite{van2008visualizing} to visualize the feature distribution of the baseline and DPGNet. As shown in Figure~\ref{fig:tsne}, the baseline model exhibits significant overlap between real and fake features, indicating that it lacks semantic distinction in the target domain. In contrast, DPGNet learns more general true/false semantics and significantly increases the separation between features of different categories. This larger separation surface DPGNet can more effectively bridge the gap between the source and target domains, thereby improving performance.

\section{Conclusion}
This work addresses the critical challenge of detecting deepfakes in realistic settings, where vast amounts of unlabeled data remain underutilized. We propose DPGNet, a novel framework that leverages text-guided alignment, curriculum-driven pseudo label generation, to exploit unlabeled deepfakes. By unifying visual and textual embeddings in a domain-invariant space, DPGNet captures generalizable features, selects informative samples to avoid overfitting. Extensive benchmarks show that DPGNet outperforms SOTA methods by a margin.